# Understanding-First AI Engineering: Purposes, Reference Architecture, and Architectural Patterns for Human Understanding

**Liming Zhu, Qinghua Lu, Ming Ding**
CSIRO, Australia
liming.zhu@csiro.au, qinghua.lu@csiro.au, ming.ding@csiro.au

## Abstract

AI systems can produce reliable results without developing human understanding. However, human understanding remains valuable for shaping subsequent AI-supported work, forming preferences, governing collective choices and satisfying curiosity, even when human participation is unnecessary for correcting errors. We propose understanding-first AI engineering, which treats purpose-specific human understanding as a first-class design principle guiding requirements, design and operation. The purposes of understanding are organised into four complementary rationales: reflective agency, effective action, collective governance and epistemic value. A reference architecture connects production, verification, use, and learning and exploration through shared records, supporting understanding through explanation and participation. Five candidate architectural patterns guide choices about parallel learning, explanatory models, goal exploration, audience-specific views and distributed expertise. Qualitative scenario analysis of two hypothetical use cases demonstrates how the purpose classification, reference architecture and selected patterns guide investigation, participation and decision processes while retaining operational controls. The cases illustrate design consequences and limitations rather than establish empirical effectiveness. The approach makes human understanding an explicit engineering commitment whose justification need not depend on improving the original AI task.

## 1 Introduction

As AI becomes more capable of producing and checking results, human understanding can cease to be a necessary input to particular tasks while remaining an important outcome of the activities surrounding them. On 6 October 2026, OpenAI released a collection of AI-generated mathematical manuscripts and supporting proof artefacts (OpenAI, 2026). Fields Medallists have cautioned that treating mathematical problem-solving as a benchmark for AI can prioritise the production of solutions over the development of human understanding (Avila et al., 2026). This concern highlights the need to support human understanding explicitly as AI becomes increasingly capable of producing results. In response to the October release, the Association for Human Mathematics calls for a "return to a vision of science that centers human understanding" (Association for Human Mathematics, 2026). This call motivates an engineering question: how can AI systems be designed so that producing results also supports humans in developing understanding?

Human involvement is often justified by its contribution to detecting and correcting mistakes. Consider instead a task for which AI produces and verifies results more reliably than additional human intervention. What purposes of understanding remain? People still need to shape subsequent AI-supported work, decide what they want, deliberate about shared priorities and make sense of the world they inhabit. If AI eventually undertakes most economically productive work and scientific discovery, these purposes could make understanding one of humanity's enduring activities. This possibility motivates the paper without making it depend on that future arriving. Understanding can be valuable because it belongs to the recipient: an AI's discovery or explanation does not itself establish a corresponding state in a person or community. The justification therefore survives the loss of a human performance advantage.

Our earlier work on verifiability-first AI engineering treats verifiability as a first-class design objective, emphasising the formulation and decomposition of problems to enable scalable and automated verification (Zhu and Lu, 2026). Human understanding can support verification and trustworthy operation, and these remain important effective-action purposes. Other purposes remain when technical verification is already sufficient: people may reconsider a goal, deliberate about an acceptable trade-off or pursue a question out of curiosity. Successful verification does not itself establish human understanding, and understanding remains valuable even when further verification does not require human understanding.

Existing research provides substantial foundations for supporting human understanding. Intellectual augmentation considers people, tools and methods together (Engelbart, 1962), while research on scientific understanding examines how AI can contribute to conceptual insight (Krenn et al., 2022). Tutoring and adaptive assistance support human learning, including approaches that explicitly optimise learning objectives in AI-assisted decision-making (Buçinca et al., 2026). Human-centred explainable AI (Ehsan and Riedl, 2020), reflective reward design (Blair et al., 2025) and meaningful human control (Cavalcante Siebert et al., 2023) address related questions concerning explanation, preference formation and responsibility. A practitioner account of iterative AI-supported development also highlights understanding as a basis for identifying possibilities and shaping subsequent tasks (Litt, 2026).

These foundations motivate an integration problem for AI engineering: how should different purposes of human understanding inform requirements, system architecture and design choices across production, verification and use? Providing explanations is one part of this problem. Systems may also need to support human participation in the activities that generate results, exploration of alternative goals and the maintenance of understanding across humans and organisations. Engineering guidance must connect these activities to explicit requirements concerning whose understanding matters, what the intended recipients should understand and why. The proposed contribution is the integration of these requirements with architectural responsibilities and pattern-selection conditions. The distinctive design question is whether an understanding requirement changes what the system investigates, represents, allocates resources to or enables people to do, including when the accepted output and deployment decision remain unchanged. An existing tutor or explanatory interface may fulfil such a requirement; additional components are justified only when needed.

This paper proposes understanding-first AI engineering, an approach that treats human understanding as an independently valuable outcome and translates that commitment into explicit requirements that guide system design, development and operation. This commitment recognises the value of understanding independently of improvements to the original AI task. “First” denotes a commitment to considering human understanding from the outset, while allowing understanding to develop during and after AI-supported activities. The main contributions of this paper are as follows:

- **Purposes of human understanding.** We classify purposes according to four complementary rationales: reflective agency, effective action, collective governance and epistemic value. We explain how these rationales inform understanding requirements.
- **Reference architecture.** We propose a reference architecture for understanding-first AI engineering, with components that support production, verification, use, and learning and exploration through shared records, enabling humans to learn through explanation and direct or parallel participation in these activities.
- **Architectural patterns.** We introduce five candidate architectural patterns to guide design choices: parallel learning workflow, separate explanatory model, branch and compare goals, shared evidence with audience-specific views, and distributed expertise with tested handover.

We examine the design consequences of the proposed approach through qualitative scenario analysis of two hypothetical use cases involving an AI-discovered energy technology and AI-supported public investment decisions. The cases illustrate how the purpose classification informs understanding requirements and how the reference architecture and selected architectural patterns guide design choices, while identifying trade-offs and limitations.

The remainder of the paper is organised as follows. Section 2 presents the conceptual background, including accountability, the value of understanding-related work and associated limits, followed by related work. Section 3 describes the methodology. Section 4 classifies the purposes of human understanding. Sections 5 and 6 present the reference architecture and architectural patterns, respectively. Section 7 presents the evaluation through two illustrative use cases and the resulting findings. Section 8 concludes the paper and outlines future work.

## 2 Background and Related Work

### 2.1 Human understanding

For an individual, understanding is an epistemic state characterised by a sufficiently accurate and connected grasp of a subject, which may concern mechanisms, mathematical relationships, constraints, possibilities, uncertainty or one's own purposes. This grasp supports relevant reasoning and revision. The adequacy of understanding depends on the recipient, the object and the purpose. A materials scientist and a community representative can each understand a technology adequately for their respective responsibilities without possessing interchangeable knowledge. Context-sensitive scientific understanding provides an important antecedent (de Regt, 2017). For engineering purposes, this definition motivates observable, purpose-specific indicators of grasp.

What counts as adequate understanding depends on the recipient, the object and the purpose, but what is true does not depend on the recipient's preferences. An explanation that a recipient finds satisfying can still be false, and a useful simplification must preserve the relationships relevant to its intended use. Similarly, an AI's possession of a sophisticated internal representation does not mean that a human recipient has developed corresponding understanding. Whether AI itself understands is a separate question. The focus of this paper is on what individuals, organisations or communities come to understand, including through appropriate tools and collaboration.

Organisations and scientific communities need not be assumed to possess a collective mind. We define collective understanding operationally as a capacity in which members have complementary understanding, can locate and combine relevant expertise, can examine disagreements, and can maintain or renew this capacity as evidence or membership changes. Distributed cognition provides a basis for examining humans, artefacts and their coordination together (Hollan et al., 2000). A repository containing an answer is a resource, but collective understanding also requires usable connections between that resource and human capabilities. No individual needs to understand everything, and consensus is not required.

Understanding also needs to be distinguished from learning. Learning can change knowledge, skills or dispositions, whereas understanding is the particular epistemic state sought here. Existing understanding may therefore be maintained without further learning gains. Producing a correct answer, expressing confidence or finding an explanation clear is insufficient, on its own, to demonstrate understanding. Learning science has long distinguished immediate performance from durable learning (Soderstrom and Bjork, 2015). Messeri and Crockett (2024) similarly caution against illusions of understanding in AI-enabled science. Assessment therefore requires evidence of understanding relevant to its intended purpose, while acknowledging uncertainty in that evidence. Understanding can also support continued participation in AI-supported work. In a practitioner account of iterative software development, Litt (2026) argues that understanding the evolving system helps humans generate ideas for subsequent development, even when checking the current output becomes less necessary.

Understanding a phenomenon differs from understanding how an AI generates outputs. Sullivan (2022) argues that humans can gain understanding of a phenomenon without knowing every detail of the model's computation. However, humans still need to understand what the model's results mean and have evidence that the results describe the phenomenon being studied. Two distinct questions therefore concern whether an explanation is supported by evidence and whether humans understand the explanation. Further investigation can help establish the evidential support for an explanation, while instruction, exploration or practice can help humans develop understanding. Addressing both needs may require further investigation together with support for human learning.

### 2.2 Accountability, value and limits

Delegating technical judgement to AI does not by itself determine who should set goals, explain institutional commitments or answer for consequences. Human and organisational accountability therefore remains relevant where decision-making arrangements assign such responsibilities, even when AI performs particular tasks more accurately. Understanding priorities, trade-offs and decision rules may be necessary beyond the verification of technical results. Future institutions may allocate responsibilities differently, so the argument does not depend on permanent human authority over every AI activity.

Accountability requires understanding appropriate to the assigned responsibility, together with access to relevant evidence, decision authority and opportunities to respond. Meaningful human control connects systems to relevant human reasons and responsible people (Santoni de Sio and van den Hoven, 2018). Assigning responsibility without sufficient control can create the moral crumple zone analysed by Elish (2019). Conversely, a lack of understanding does not necessarily remove responsibility when an organisation has an obligation to develop or obtain the required knowledge. Supporting understanding cannot by itself establish legitimate authority or settle legal liability. The adequacy of understanding remains dependent on the recipient, the object and the purpose.

Accountability is one justification for supporting understanding, rather than a necessary condition for valuing understanding. Curiosity, self-understanding and participation in a shared intellectual culture may remain valued even when human responsibility for particular AI activities is reduced. In the future considered in the introduction, human understanding could become an explicit purpose of otherwise autonomous work. Such value depends on what recipients and communities seek to understand, rather than on maintaining a performance advantage over AI. Appropriate targets may include a partial but accurate grasp, recognition of dependence on specialist expertise and awareness of unresolved questions. Full comprehension of every discovery is not required. These purposes do not merely provide humans with activity after automation. Understanding alternatives helps people form and express preferences; understanding consequences and others' reasons supports collective self-government. Curiosity can motivate extending a shared body of intelligible knowledge. Recognising the boundary between what can be explained, what depends on others' expertise and what remains unknown also helps recipients frame questions, collaborate and judge the scope of their reliance. These boundaries are provisional and may be extended through inquiry and learning.

The value of understanding also has implications for how understanding-related work is supported and rewarded. Professional status, research funding or paid roles may support activities that make discoveries comprehensible or maintain collective expertise, even without improving the accuracy or output of the original AI task. Such arrangements would recognise the value of developing and sustaining understanding. Economic support for these activities would depend on collective priorities, budgets and incentives. Such support is an institutional possibility rather than a prediction of future employment or labour-market outcomes.

Supporting understanding also raises risks. Assessment may confuse rehearsed responses with understanding, personalisation may become persuasion or surveillance, and shared accounts may suppress disagreement. Unequal access to time, education and expert support may also restrict participation. Consequently, understanding requirements should remain open to revision, assessment should acknowledge uncertainty, and system design should preserve disagreement and informed choices to stop or delegate. Investigation, learning and coordination also consume resources, so greater understanding cannot be pursued without regard to cost or consent.

### 2.3 Related work

Research on AI-supported understanding examines how humans gain insight from computational results. Krenn et al. (2022) discuss AI's role in developing scientific understanding, while Ehsan and Riedl (2020) emphasise the recipient and social context of explanations. Recent software-engineering studies investigate explanations for vulnerability predictions (Nguyen et al., 2025) and compare explanation strategies in smart environments (Sadeghi et al., 2026). In our study, we consider both explanation and human participation in production, verification and use as ways to develop understanding.

Human learning is an explicit objective in adaptive assistance. Buçinca et al. (2026) optimise decision support for both decision accuracy and learning, finding that learning improvements are less consistent. Bastani et al. (2025) report reduced unassisted performance following unrestricted AI assistance, while Kestin et al. (2025) demonstrate learning benefits from a structured AI tutor. These findings highlight the importance of assistance design and of distinguishing successful task completion from human learning. These studies concern particular populations, tasks and forms of assistance. However, their findings do not establish that AI assistance generally improves or harms understanding.

Other research addresses reflection, responsibility and collaboration. Blair et al. (2025) use reflective dialogues to support preference construction. Cavalcante Siebert et al. (2023) identify system properties for meaningful human control, and Clemmensen et al. (2025) review socio-technical perspectives on human-in-the-loop systems. Lestingi et al. (2026) combine explanation, adaptation and formal verification in human–machine teaming. In our study, we classify the purposes of human understanding and propose a reference architecture and architectural patterns to support those purposes in AI systems. The focus is on relating purpose-specific understanding requirements to components, their relationships and pattern selection across production, verification and use. Existing explanatory systems or tutors may already provide the required support in particular applications. The proposed integration gives these requirements consequences across investigation, representation, participation and renewal, rather than treating provision of an explanation as sufficient evidence that a recipient understands.

## 3 Methodology

Figure 1 summarises the development of the approach through narrative literature synthesis, insights from previous projects and design reasoning. The purposes of understanding establish why understanding matters and help specify what recipients need to understand. The reference architecture identifies supporting responsibilities and relationships, and candidate patterns describe recurring design choices. Two hypothetical use cases examine how these elements work together, including adverse conditions and simpler alternatives. The prior-project insights draw on the authors' experience in AI-for-science and industry research projects. The scenarios provide analytical illustrations rather than observations of deployed systems.

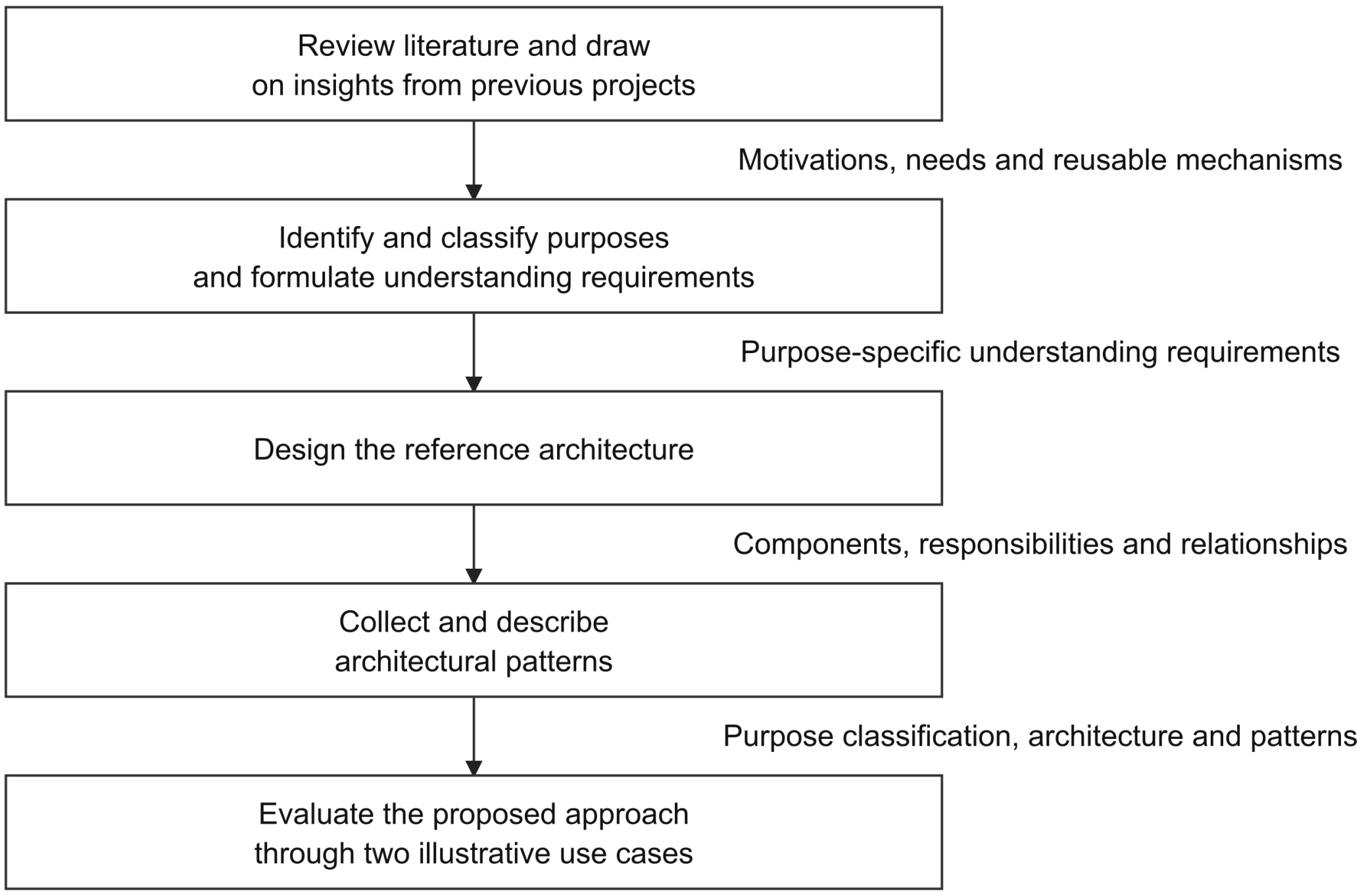

Figure 1. Methodology.

The literature review brings together motivations for human understanding, conditions for adequate understanding, and mechanisms for supporting understanding in AI systems. Relevant concepts and findings are examined for the recipients and purposes addressed, the requirements implied, and the mechanisms and limitations described. Sections 2 and 4 identify the conceptual foundations, while the documented applications in Section 6 identify reusable mechanisms and the limits of their correspondence to the proposed patterns. The source set combines conceptual accounts, empirical studies and documented mechanisms relevant to the design problem. Their evidential roles differ: conceptual accounts motivate requirements, empirical studies support claims only within their studied settings, and practitioner accounts and product documentation illustrate mechanisms. The synthesis is selective and does not use a systematic search and screening protocol. It therefore supports a reasoned proposal, without establishing literature completeness, an exhaustive classification or empirical validation of the resulting designs.

We identify the purposes of human understanding by bringing together motivations discussed in the literature and reasoning about recurring design problems. We group these motivations according to the value that understanding provides, resulting in the four complementary rationales presented in Section 4. For a particular application, the rationale is considered together with the recipient, object, required scope and depth, permitted support and constraints to formulate understanding requirements. Multiple rationales may inform the same requirement, and a rationale does not prescribe a particular pattern. The synthesis does not constitute an independent empirical evaluation.

Three source-to-design links illustrate the synthesis. Context-sensitive understanding (de Regt, 2017) motivates specifying the recipient, object and purpose in each requirement. The distinction between learning and performance (Soderstrom and Bjork, 2015) motivates separate evidence for output acceptance and recipient understanding. Distributed cognition (Hollan et al., 2000) motivates treating access to complementary expertise and coordination as responsibilities beyond storing information. These are design interpretations of the cited work. They inform, respectively, the requirement description in Section 4, the separate records and feedback routes in Section 5, and the audience coordination and handover mechanisms in Section 6. This mapping explains the proposed design rationale retrospectively, rather than claiming that the sources establish the complete architecture or pattern catalogue.

The reference architecture is developed with guidance from Galster and Avgeriou (2011). We classify the proposed architecture as an industry-cross-cutting, preliminary facilitation architecture intended for use across organisations, drawing on the classification of Angelov et al. (2009). Understanding requirements guide the identification of responsibilities for production, verification, use, learning and exploration, and management of shared records. These responsibilities are assigned to logical components, with relationships defined by the exchange of candidate outputs, claims, records, requests and feedback. The resulting architecture provides design guidance across application domains without requiring a separate service for every component. The illustrative use cases examine how the components and relationships can support particular requirements.

We propose candidate architectural patterns by adapting mechanisms from the literature and engineering practice to recurring design problems. The pattern template comprises pattern name, context, problem, forces, solution, consequences and known uses, drawing on the pattern-writing guidance collected by Martin et al. (1997). Known uses identify documented applications of the underlying mechanisms and clarify the extent to which those applications correspond to the proposed patterns. Benefits and drawbacks describe anticipated design consequences. The illustrative use cases examine the application of selected patterns with the reference architecture.

We examine the approach through qualitative scenario analysis of an AI-discovered energy technology and AI-supported public investment decisions. Each case identifies the recipients, objects and purposes of understanding, specifies the associated requirements, instantiates the reference architecture and selects relevant patterns. We trace scenarios through the designs, examine adverse conditions and compare plausible alternative designs. Requirements are considered addressed when the design specifies relevant responsibilities, supporting records and relationships, partially addressed when additional design or institutional arrangements remain necessary, and unresolved when the analysis provides no adequate support. These judgements concern the proposed designs rather than measured attainment of human understanding. Section 7 reports the findings for the purpose classification, reference architecture and architectural patterns. The cases contrast understanding of a technical phenomenon under stable operational goals with reflection on evolving goals in a collective decision. They are selected to expose different design consequences, not to force every pattern into each system.

Three questions organise the analysis: which understanding requirements follow from the selected purposes, which architectural responsibilities and records support those requirements, and when selected patterns are justified compared with simpler arrangements? The unit of analysis is a requirement and its supporting design decisions within a scenario. For each comparison, the relevant recipients, information and operational constraints are held fixed by assumption. We examine whether the design identifies an activity, responsible role, supporting record and follow-up route, and what remains unspecified if one of these is absent. These criteria concern traceability and completeness of the proposed arrangements within the stated scenario, not measured learning or comparative effectiveness.

## 4 Purposes of Human Understanding

Understanding can be valuable as a means to another end or as an end in itself. Within the first branch, we distinguish determining what to pursue, acting effectively, and governing collective choices. The second branch concerns the value of possessing understanding. Table 1 summarises these purposes through four complementary rationales, namely reflective agency, effective action, collective governance, and epistemic value. The distinction concerns why understanding is wanted and does not imply a sequence of cognitive stages or a ranking of worth. A single activity may serve several rationales, and epistemic value can accompany any instrumental benefit.

Table 1. Purposes of human understanding, why understanding matters, illustrative examples and system support.

| Purpose of understanding | Why understanding matters | Illustrative example | How the system supports understanding |
|---|---|---|---|
| Reflective agency | Helps humans determine what to pursue and understand why their preferences may change. | A person understands the implications of alternative career paths and reconsiders which goals to pursue. | Supports self-understanding and preference formation through exploration and comparison of alternatives, allowing goals to be revised before changes are put into effect. |
| Effective action | Helps humans achieve an accepted goal, detect faults and improve results. | An engineer understands why an AI-generated program fails so that the fault can be corrected. | Provides explanations, relevant evidence and tools to understand results and limitations, investigate faults and check repairs, supporting application, improvement and warranted reliance. |
| Collective governance | Helps humans understand who has authority to decide, how trade-offs are handled and who is accountable for decisions. | Community representatives understand who benefits from a public investment proposal and who bears the risks. | Makes evidence and trade-offs accessible to participants in different roles, records disagreements and reasons for decisions, and supports legitimate processes for revising decisions. |
| Epistemic value | Enables humans to pursue understanding as an end in itself. | A person studies an AI-generated mathematical proof out of curiosity. | Enables explanation, exploration and learning without requiring a practical benefit. |

Reflective agency concerns the goals that an individual or group chooses to pursue. Understanding alternatives may lead to changes in those goals, rather than improvements against an existing goal, as discussed by Carenini and Poole (2002) and Blair et al. (2025). Systems can support exploration and comparison of alternatives, allowing goals to be revised before changes are put into effect. In iterative AI-supported development, understanding the evolving system can help humans identify new possibilities and formulate subsequent tasks (Litt, 2026). This supports reflective agency when understanding informs what to pursue next, even when the current AI-generated result requires no correction.

Effective action includes understanding results and limitations to support fault detection, diagnosis, correction, improvement, appropriate reliance and transfer to new situations. Explanations can help humans assess model predictions and decide when to rely on them (Ribeiro et al., 2016). A developer may need to understand a failure to correct the underlying fault, while an operator may need to understand a limitation to prevent unsafe use. These applications are central to the proposed approach, which supports understanding through explanations, relevant evidence and tools to investigate faults and check repairs. Removing the need for human correction is a limiting case that makes the other rationales visible; it does not remove the need to form preferences, govern collective choices or pursue curiosity. The approach does not exclude applications in which human understanding improves system performance.

Collective governance concerns decision authority, the representation of interests and trade-offs, and institutional accountability (Santoni de Sio and van den Hoven, 2018). Systems can support these purposes by making evidence and trade-offs accessible to participants in different roles, recording disagreements and reasons for decisions, and supporting legitimate processes for revising decisions. Although these rationales overlap, identifying an individual preference does not establish the legitimacy of a collective decision. Democratic participation may require understanding other participants' reasons and institutional choices, even when individual preferences are settled. Understanding alone does not resolve value conflicts or legitimise decisions.

Epistemic value concerns understanding as a valued state, including curiosity and participation in an intellectual culture (Pritchard, 2009). Curiosity may concern a general gap in knowledge or arise from a particular encounter, experience or situation. These motivations can overlap: situated curiosity can initiate a broader epistemic inquiry. AI-supported exploration may also contribute to scientific understanding (Krenn et al., 2022). For epistemic value, systems can enable explanation, exploration and learning without requiring a practical benefit. Learning describes changes in that state rather than a separate purpose. Investigation may support practical improvement or curiosity, depending on the motivation. Similarly, maintaining collective knowledge concerns the distribution and continuity of understanding and may serve all four rationales. Learning, investigation and knowledge maintenance are therefore activities that support the purposes of understanding, rather than additional rationales.

The rationale for understanding should be distinguished from its recipient and object, and from the activities through which understanding is acquired or maintained. A scientific community may seek understanding to support effective action or for its own sake, while an individual may seek understanding for public accountability. The proposed rationales provide an extensible classification rather than an exhaustive taxonomy. Their role is to distinguish understanding requirements while allowing multiple rationales to apply to the same case.

An understanding requirement specifies the relevant rationale, recipient, object, required scope and depth, permitted support, evidence of attainment where assessment is appropriate, resource budget and revision conditions. The requirement also specifies who is authorised to revise the requirement. For reflective agency, an individual may need to explain how exploring a prototype changed their priorities. For debugging, an engineer may need to localise a fault and justify a repair using logs and tests. For collective governance, a committee may need to distinguish technical evidence from contested policy choices. A requirement motivated by epistemic value may permit open exploration without formal assessment or operational responsibility. For an assessed requirement, specify the task, allowed tools, assessment criteria, assessor and renewal trigger. For example, an operator may use approved records to interpret an unfamiliar operating condition, identify a relevant limit and justify escalation; the responsible assessor agrees the criteria before the exercise.

Understanding requirements may conflict. An explanation sufficient for immediate correction may omit relationships relevant to curiosity. A shared organisational explanation may facilitate coordination while suppressing disagreement relevant to public deliberation. Requirements should therefore specify the selected rationales, intended recipients and acceptable trade-offs. A system need not support every rationale or recipient, and delegation may be appropriate within any configuration.

## 5 Reference Architecture for Understanding-First AI Engineering

### 5.1 Conventional, verifiability-first and understanding-first design

A conventional task workflow provides a starting point: (1) requirements specify what is wanted, (2) production creates a candidate, (3) verification assesses relevant claims, and (4) accepted results are used or deployed. This describes a simplified workflow rather than a rigid sequence of activities. Each activity may involve AI, humans or both. Verification may use automated checks, the producing AI, other systems or human judgement, with independence and rigour appropriate to the claim. Failed checks return work for correction, while experience in use may revise requirements or trigger investigation. Requirements are recorded specifications that guide activities, rather than a workflow activity of the same kind as production, verification and use.

Verifiability-first AI engineering makes the ability to check claims a design constraint from the outset (Zhu and Lu, 2026). The approach can change the problem decomposition, intermediate artefacts and allocation of checking across the workflow, rather than merely add a final test (Lu et al., 2026). Understanding-first AI engineering applies the same design-time logic to selected human understanding requirements. The approach asks what humans should be able to grasp, why, and how the workflow must change to make that understanding attainable. Verifiability-first AI engineering and understanding-first AI engineering can therefore be used together.

“First” denotes priority in problem formulation and architectural decisions, not temporal precedence or unconditional dominance. Human comprehension need not precede every action, and understanding does not override safety, privacy, cost or other requirements. The term emphasises making understanding-related design commitments early enough to affect the architecture.  A system is understanding-first when those commitments can affect its design before explanations are produced, even if the resulting learning occurs after production.

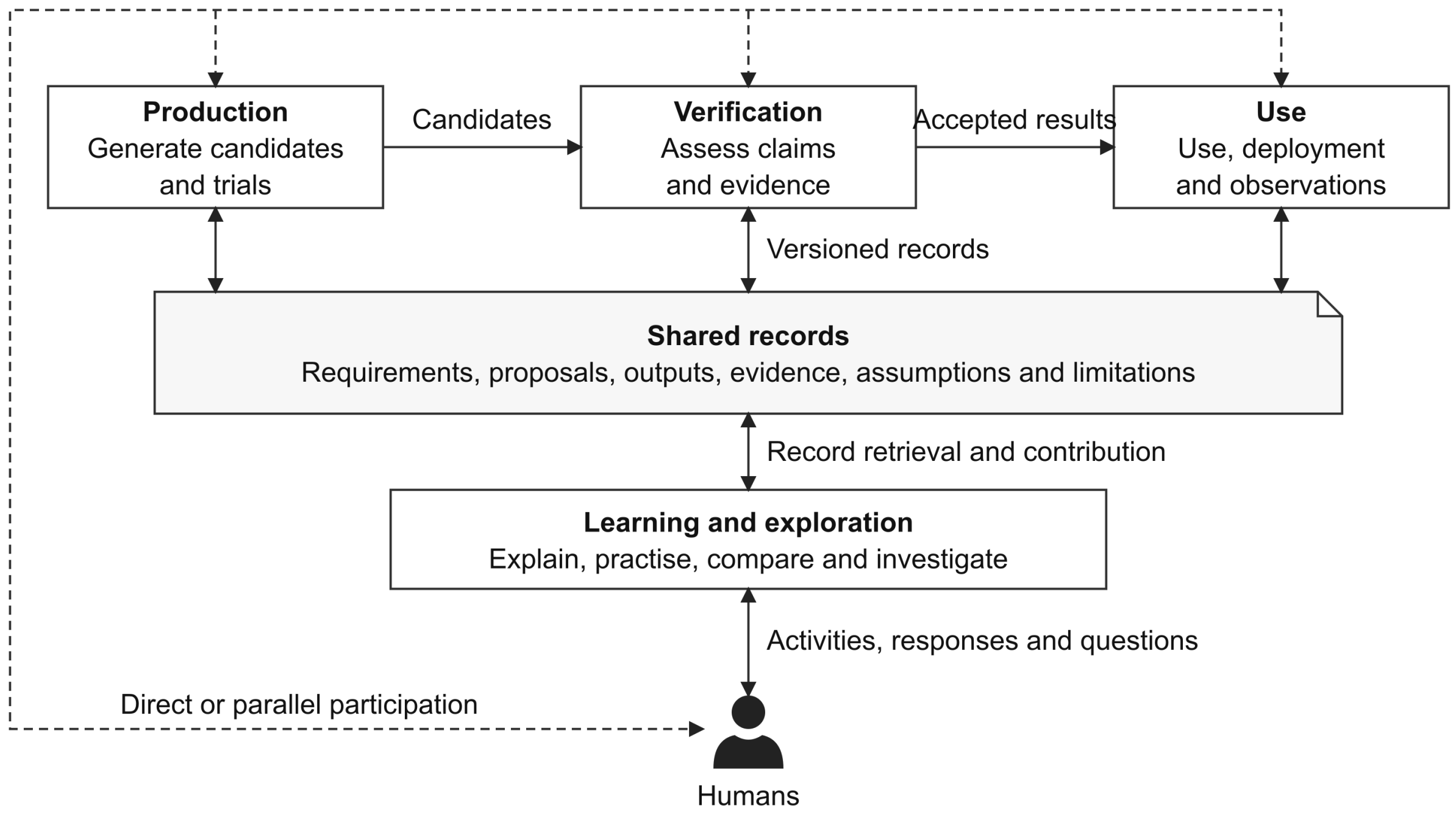

Figure 2. Reference Architecture for Understanding-First AI Engineering.

### 5.2 Reference architecture design

Figure 2 presents the proposed reference architecture for understanding-first AI engineering, showing the logical software components, their responsibilities and relationships. Production generates candidate outputs and trials. Verification assesses relevant claims and evidence. The use component supports deployment and use and records operational observations. The learning and exploration component provides explanation, practice, comparison and investigation. The shared records component maintains requirements and their revisions, candidate outputs, attempted approaches, test results, operational observations, evidence, assumptions and limitations. Production can reveal informative dead ends, verification can expose a boundary, and use can reveal an unanticipated interaction. Learning and exploration also contribute questions, exploratory artefacts, reflections and, where appropriate, evidence of attainment. Records retain provenance and status because a recorded claim or plausible explanation is not automatically established evidence.

The components define logical responsibilities rather than mandatory deployment units. A component may use several services, and one service may support several components. The shared records component can connect existing repositories, logs, models and decision records while preserving access limits and distinguishing approved requirements from exploratory proposals. Production passes candidates to verification, and verification passes accepted results to use. The components read and contribute shared records, supporting exchanges of verification findings, operational observations, investigation requests and proposed revisions. The learning and exploration component retrieves and contributes records and provides access to the activities that generate experience. Humans participate individually or collectively as users and recipients of the supported activities. Software supports human and AI activities without representing human understanding as a stored system state. Paired arrows indicate record access or interaction, and the dashed route illustrates direct or parallel human participation in production, verification and use. These relationships support feedback and revision rather than a rigid execution sequence. Software components support these responsibilities, while named human or organisational roles own approval, assessment and review decisions. Access to a shared record does not itself confer authority to change an operational goal or accept a claim.

Understanding can develop through doing. A person can construct an example, reproduce a result, formulate and execute checks, inspect a failure, or try an alternative use. The architecture therefore permits human production and verification alongside AI work, with or without AI assistance. Such activity may use live work, a recorded case, a simulation or a parallel workspace. The value of participation can lie in developing the participant's grasp without adding new output or checking capacity to the main workflow. Participation is not proof of understanding: the activity should expose relevant relationships and prompt interpretation, comparison or revision. Likewise, a learning exercise counts as operational verification only if the exercise independently meets the applicable checking requirements.

The learning and exploration component can estimate existing grasp, select activities and examine reasoning, drawing on learner modelling and adaptive assistance. An evidential gap may call for investigation. A gap in grasp may call for explanation or practice. Good teaching cannot settle an unsupported mechanism. For an organisation, assessment also concerns access to complementary expertise and the ability to combine that expertise. Estimates should remain correctable and proportionate to purpose. Voluntary curiosity does not imply permission for competence surveillance. Where assessment is appropriate, the learning loop relates a selected requirement to an activity, evidence of the recipient's reasoning and subsequent adaptation or renewal. A completed activity or accessible record alone does not close that loop.

### 5.3 Feedback with different destinations

The four rationales help determine where feedback goes. Reflective agency can propose a change to goals or constraints. Effective action can send a fault hypothesis or proposed correction to production and verification. Collective governance can trigger deliberation, escalation or a review of the authority under which a decision was made. Epistemic value can justify further exploration without changing the deployed result. The same interaction may generate more than one kind of feedback, but each needs an identified destination.

These routes have different control conditions. A proposed repair is checked before use. A proposed goal revision enters the appropriate personal or institutional decision process. Learning or a changed recorded preference does not itself authorise a deployed goal change. Parallel work can have its own budget and completion state, and may repeat already successful production or verification solely to develop understanding. Where live participation creates unacceptable operational risk, replay or simulation supplies a separate setting. A material defect found in any setting enters the ordinary correction process. An activity that only develops understanding can finish without changing production.

For iterative development, we propose linking learning outcomes to subsequent design decisions through the shared records. Questions and proposed changes from learning and exploration can become inputs to production, verification or goal review, according to their purpose. Records should identify the system version understood and the decision that the understanding supports. This connects the learning loop to system evolution without treating a learning result as authorisation to modify the system.

The implementation should retain separate records of output acceptance, evidence of recipient understanding and approved changes to goals or responsibilities. An understanding record states its scope, evidence, date, uncertainty and allowed support. The record cannot become a universal credential. Renewal may follow changed evidence, a new responsibility or staff turnover even when software behaviour remains unchanged. The architecture assigns responsibilities to components and makes their relationships explicit while leaving implementation, timing and distribution to application-specific design choices.

## 6 Architectural Patterns for Human Understanding

A reference architecture identifies components, their responsibilities and relationships across a system, while a pattern addresses a recurring problem under particular conditions. In this section, we discuss five candidate architectural patterns that guide the design of learning workflows, explanatory models, goal exploration, audience-specific communication and knowledge handover. The patterns adapt established mechanisms to understanding-first AI systems and can be used individually or combined according to the purposes and requirements of human understanding. The benefits and drawbacks describe anticipated consequences of these design choices:

- P1 separates the timing of learning from operational work;
- P2 separates explanatory and production representations;
- P3 separates exploration from commitment;
- P4 coordinates views across audiences; and
- P5 preserves access to expertise across changes in responsibility.

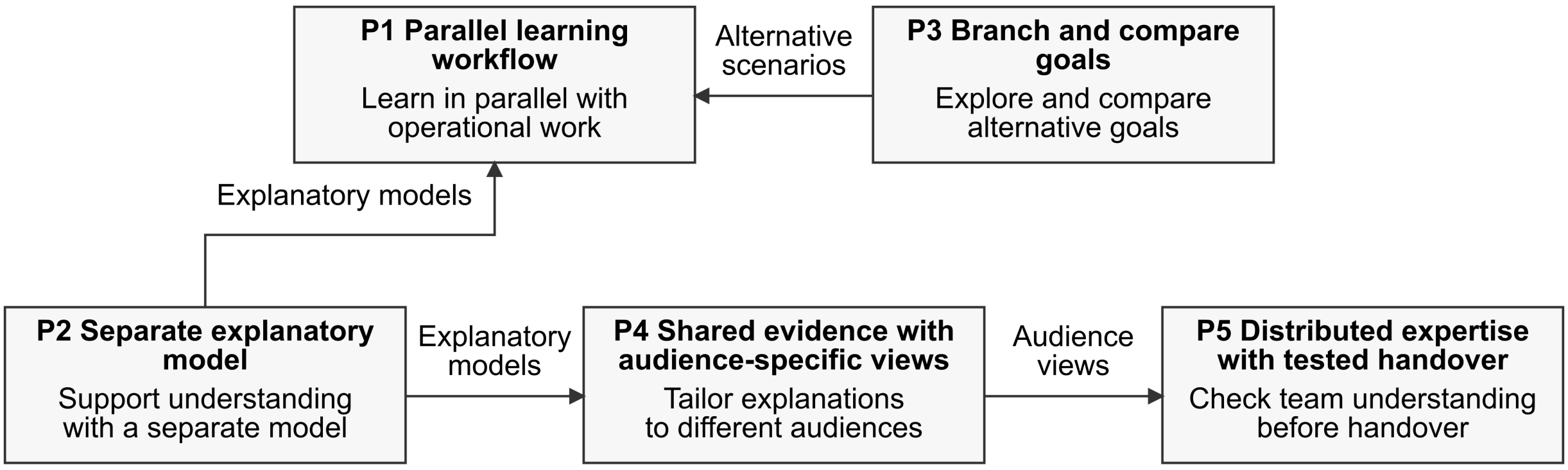

Figure 3. Pattern relations.

Figure 3 shows how the patterns can support one another. Branch and compare goals (P3) can provide alternative scenarios for comparison in a parallel learning workflow (P1). A separate explanatory model (P2) can support learning activities in P1 and provide the basis for audience-specific explanations in P4. Those explanations can help receiving teams prepare for the handover checks in P5. The arrows identify optional inputs and do not prescribe a sequence or require all five patterns to be combined.

The known-use examples identify documented applications of underlying mechanisms, rather than evidence that the complete candidate patterns have been validated. Each description identifies the supported mechanism and any material differences from the proposed pattern. Section 7 illustrates application of selected patterns within understanding-first AI systems. In Figures 4–8, rectangles represent activities or model components, document shapes represent information artefacts, and human icons identify participants.

### 6.1 Parallel learning workflow (P1)

**Context.** Autonomous AI work can finish faster than humans can acquire the desired understanding. Human participation may be needed for training in fault diagnosis and correction, preparation for accountable roles or curiosity, although the urgency differs.

**Problem.** How can an AI system support human understanding without making every production step wait for learning?

**Forces.**

- **Timeliness.** Production may need to finish promptly, while human learning requires time and practice.
- **Safety.** Learning requires exploration and mistakes, while live operations must remain protected.
- **Relevance.** Learning activities need current operational evidence, although production and learning progress at different rates.

**Solution.** Provide a learning workflow alongside the productive workflow, as shown in Figure 4. Production, verification and use or deployment contribute outputs, evidence and traces to shared records. These records support practice, replay and simulation in the parallel learning workflow. Humans can reconstruct an output, build and run checks, compare alternative solutions, or explore recorded or simulated behaviour. Checks of relevant understanding assess human responses and provide feedback against the selected understanding requirements. Give the workflows separate completion states, schedules and budgets. Human learning can continue after the AI has completed and checked the productive task. For a hypothetical automated service repair, a verified fix can be deployed while an engineer reconstructs the fault and tests candidate repairs in a parallel environment. Where an immediate human decision requires the relevant understanding, learning remains a prerequisite for that decision. For continued participation in development, set a learning checkpoint before the next decision that requires the recipient's understanding. Anchor the checkpoint to the relevant system version and assess reasoning about a possible change. Use the result to adapt learning or seek additional expertise. This need not delay unrelated autonomous work.

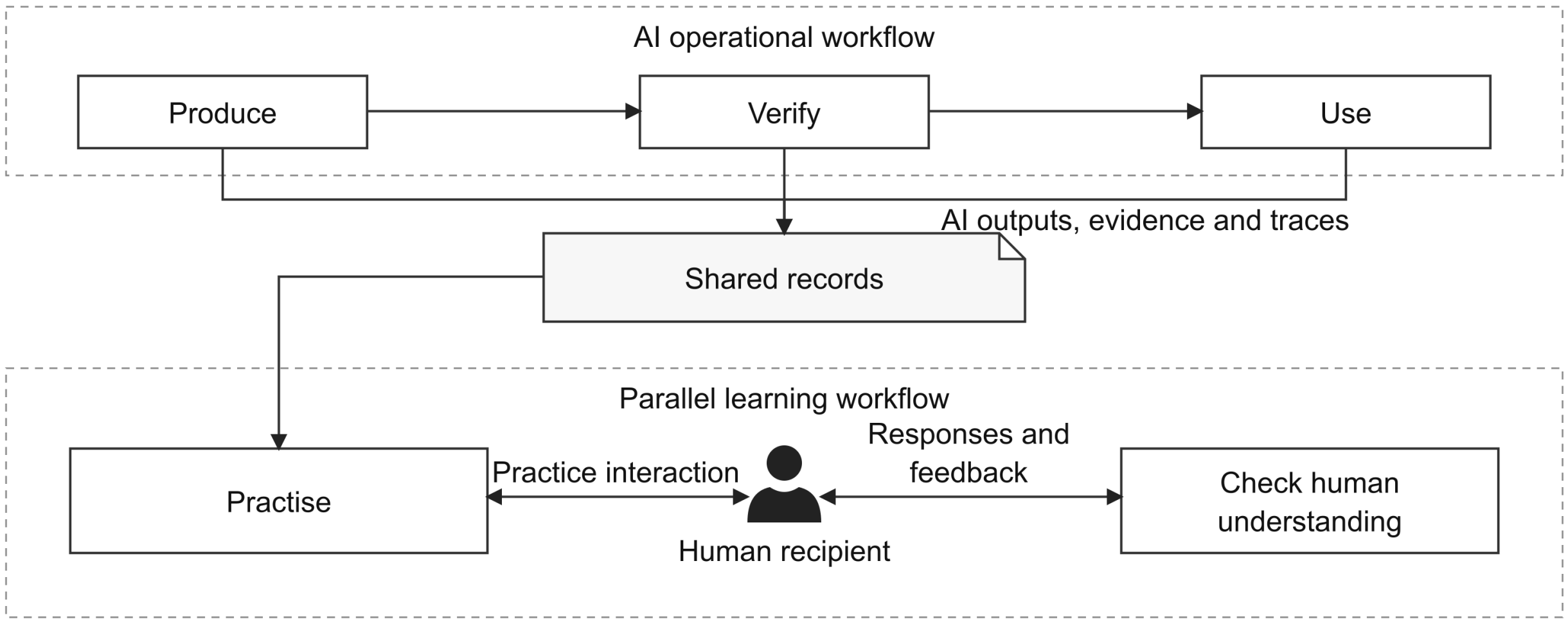

Figure 4. Parallel learning workflow.

**Consequences.**

**Benefits.**

- **Responsiveness.** Separate schedules allow production to proceed without waiting for learning unless understanding is required for an immediate decision.
- **Safety.** Practice through replay and simulation in a separate learning environment avoids exposing live operations to learning mistakes.
- **Flexibility.** Learning can support correction, preparation for accountable roles or epistemic value, without requiring a productivity gain.

**Drawbacks.**

- **Learning delay.** Production priorities can postpone learning indefinitely. Required learning needs an owner and explicit completion criteria.
- **Maintenance cost.** Exercises and simulations can become outdated as operational systems change, requiring renewal triggers and resources.
- **Additional complexity.** Separate workflows introduce scheduling and coordination overhead. Embedded coaching may suffice for brief participation, and a separate workflow is unnecessary when no understanding requirement has been selected.

**Known uses.**

- **Salesforce Agentforce Sales Coach**[1]. A coaching agent uses the context of a sales opportunity for repeatable pitch practice and negotiation role-play. Humans can practise and review feedback separately from the customer interaction, providing a concrete example of practice connected to work records.
- **Yoodli**[2]. Roleplay Agent creates practice scenarios from reference material such as real calls and presentations, with personas and rubric goals. Humans rehearse with AI personas in a separate practice environment, linking learning activities to material from operational work. These examples illustrate practice separated from customer-facing work; they do not by themselves establish the separate completion criteria, renewal controls or understanding outcomes proposed in P1.
- **Agent-generated learning tools.**[3] Litt describes code explainers with quizzes and interactive tools for exploring interpreter execution and website migration. These practitioner examples illustrate agent-supported reconstruction and exploration relevant to P1. They do not establish measured learning gains or adoption of the complete pattern. A quiz result alone is insufficient evidence of readiness for a design decision.

### 6.2 Separate explanatory model (P2)

**Context.** An AI system's production model or search process may be too complex for the intended recipient. Restricting production to an easily understood method may compromise the task.

**Problem.** How can an AI system provide an accessible representation without requiring the production model itself to be simple?

**Forces.**

- **Performance and accessibility.** Task performance may favour complex production models, while recipients need representations suited to their knowledge and purpose.

[1] https://trailhead.salesforce.com/content/learn/projects/build-your-first-sales-coach/configure-and-activate-your-sales-coaching-agent
[2] https://support.yoodli.ai/en/articles/15960287-create-roleplays-with-roleplay-agent
[3] https://www.geoffreylitt.com/2026/07/02/understanding-is-the-new-bottleneck.html

- **Fidelity.** Simplification must preserve relevant relationships without presenting unsupported explanations as established knowledge.
- **Maintainability.** Changes to the production model or supporting evidence require corresponding changes to the explanatory model.

**Solution.** Maintain a simpler explanatory model alongside the production model, with explicit scope, assumptions and links to supporting evidence, as shown in Figure 5. The production model generates outputs for use or deployment. Humans query the explanatory model and explore predictions, counterfactuals and relevant relationships through the explanations provided. Validate the explanatory model against the intended claims and renew the model when those claims change. For explanations of model behaviour, compare predictions from the explanatory and production models within the stated domain. Figure 5 shows the comparison as a separate activity that records agreement and differences. For explanations of the world, require independent evidence connecting the explanatory model to the phenomenon. Agreement with production outputs alone does not establish a scientific mechanism. For model-behaviour explanations, define the evaluation domain and acceptable disagreement, including relevant slices and boundary cases; disclose cases in which the explanatory model is unreliable.

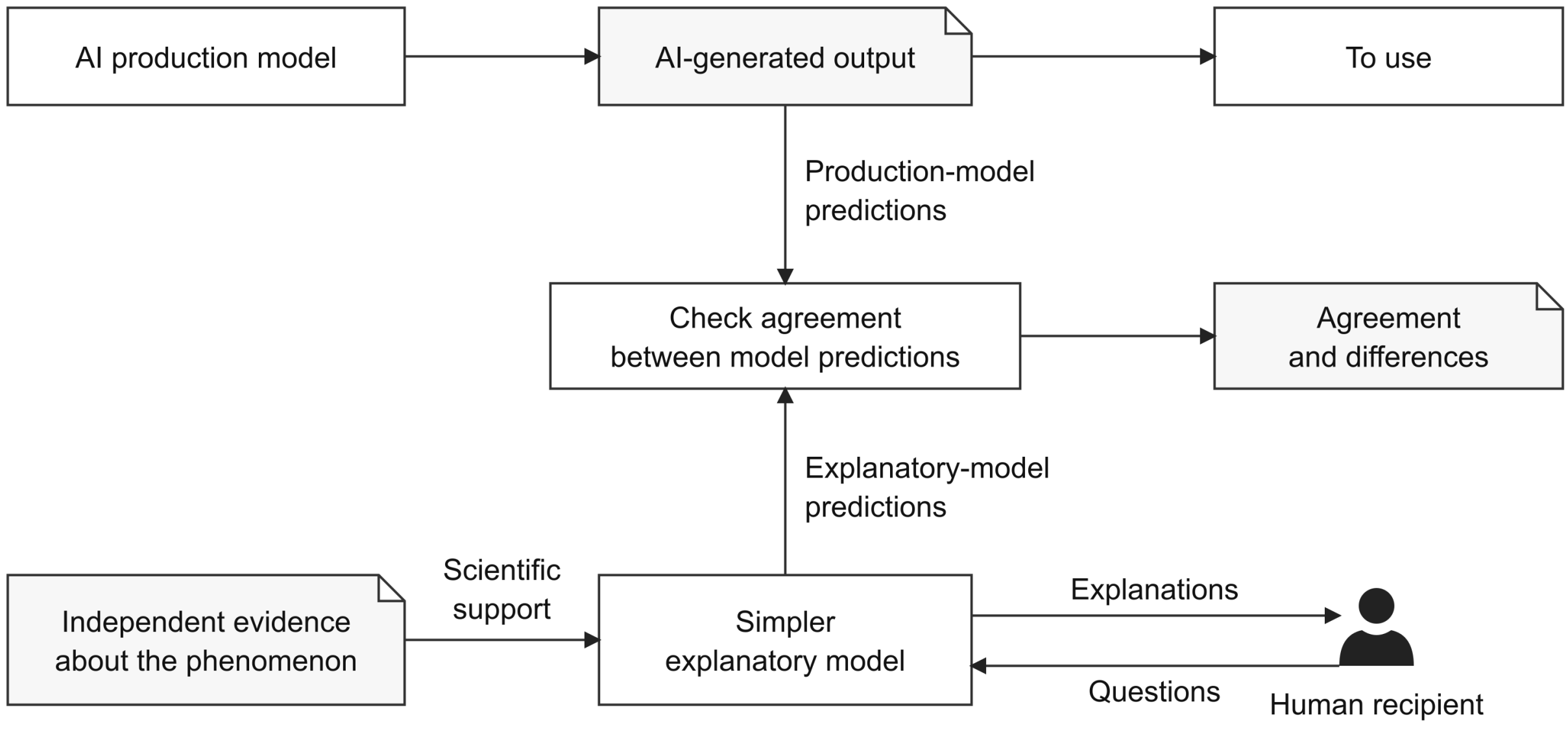

Figure 5. Separate explanatory model.

**Consequences.**

**Benefits.**

- **Flexibility.** Production methods can be selected for task requirements while explanatory representations address understanding requirements.
- **Accessibility.** A simpler representation supports prediction exercises, exploration and interpretation for the intended recipient.
- **Adaptability.** Explanatory models can address different recipients and purposes across the four rationales.

**Drawbacks.**

- **Validation cost.** The additional model requires fidelity checks and, for claims about the world, independent supporting evidence.
- **Maintenance cost.** Production and explanatory models can diverge unless changes trigger review and synchronisation.

- **Misleading simplification.** An accessible account may be mistaken for a causal mechanism. An inherently interpretable production model is preferable when both sets of requirements can be met, while direct inspection may suffice for manageable representations.

**Known uses.**

- **LIME[4].** The implementation fits a local interpretable model to explain a prediction from a separate classifier. The explanation is limited to the neighbourhood of the prediction.
- **Azure Machine Learning MimicExplainer[5].** MimicExplainer trains an interpretable surrogate to approximate the predictions of a separate model, providing a global explanation of model behaviour.
- **H2O Driverless AI[6].** The model interpretation tools provide decision-tree surrogate representations of more complex models, including exportable rules.

### 6.3 Branch and compare goals (P3)

**Context.** Humans may clarify what to pursue by exploring the consequences of alternative goals, while an AI system needs a sufficiently stable, authorised goal to guide operational behaviour. The problem is especially relevant to reflective agency and collective governance.

**Problem.** How can humans explore and revise goals without exploratory preferences silently changing operational behaviour?

**Forces.**

- **Reflection and stability.** Recipients need opportunities to revise goals, while operational behaviour requires sufficiently stable commitments.
- **Exploration cost and framing.** Comparing alternatives takes time, and the selection and presentation of alternatives can influence preferences.
- **Decision authority.** Exploration should remain open, while adoption of revised goals must follow the applicable decision process.

**Solution.** Create exploratory branches from the current authorised goal, as shown in Figure 6. The alternatives represent different goals or priorities. Use prototypes or simulations to explore the consequences of pursuing each alternative goal. Compare the predicted consequences and trade-offs to help humans decide whether to retain or revise the current goal, and record reasons for any proposed revision. Keep the branches separate from deployed configuration. Participants may reject all alternatives, and agreement with an inferred preference model does not establish human self-understanding. Authorise a chosen revision through the relevant decision process, verify applicable constraints and adopt the approved goal. For a community investment decision, participants can explore the consequences of prioritising energy output or land preservation before deciding whether to revise the investment goal. A conversational preference update does not silently change the selected project.

---

[4] https://github.com/marcotcr/lime
[5] https://learn.microsoft.com/en-gb/azure/machine-learning/how-to-machine-learning-interpretability?view=azureml-api-2
[6] https://docs.h2o.ai/driverless-ai/latest-lts/docs/userguide/interpreting.html

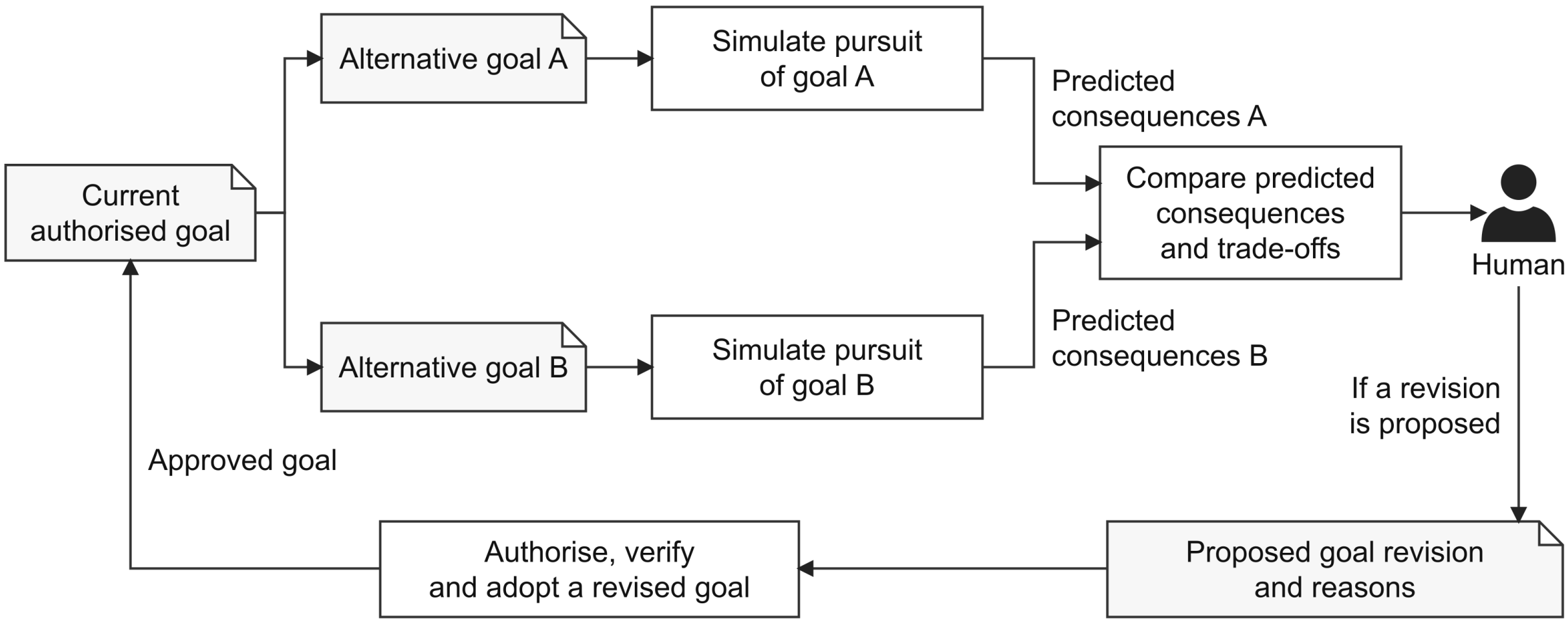

Figure 6. Branch and compare goals.

**Consequences.**

**Benefits.**

- **Reflective agency.** Comparing consequences supports preference formation before commitment.
- **Operational stability.** Exploratory changes remain separate from deployed goals until an authorised revision is adopted.
- **Traceability.** Recorded alternatives and reasons make goal revisions available for subsequent review and collective deliberation.

**Drawbacks.**

- **Decision overhead.** Branch creation and comparison require time and resources. Direct preference elicitation is simpler when goals are stable and well understood.
- **Framing bias.** Presented alternatives can restrict reflection. Recipients need contrasting framings and the ability to reject all offered options.
- **Reduced responsiveness.** Separating exploration from adoption can delay changes. Immediate adaptation may suit low-impact, reversible choices whose consequences are understood.

**Known uses.**

- **Google What-If Tool**[7]. Users can vary the relative costs of false positives and false negatives and compare the resulting classification thresholds and performance trade-offs. The application supports exploration of alternative decision priorities, although the documentation does not establish the full process of recording reasons for revised goals and authorising changes.
- **LangGraph.**[8] Developers can fork a saved agent state and resume execution along an alternative path while preserving the original history. This provides a branching mechanism for exploring changed goals encoded in state. Goal comparison and approval remain application-specific, and replay can repeat external actions.

### 6.4 Shared evidence with audience-specific views (P4)

**Context.** Several recipients need different depths and forms of understanding about the same AI-supported work. The need can arise within any rationale and is especially consequential in accountability and public deliberation.

[7] https://pair-code.github.io/what-if-tool/learn/tutorials/cost-ratio/
[8] https://docs.langchain.com/oss/python/langgraph/use-time-travel

**Problem.** How can an AI system provide explanations suited to different audiences while retaining consistent links to evidence and assumptions?

**Forces.**

- **Accessibility and consistency.** Recipients need different levels and forms of explanation, while the accounts must retain a consistent evidential basis.
- **Plurality.** Shared evidence should support comparison without suppressing legitimate disagreements or alternative interpretations.
- **Privacy.** Access to evidence must respect privacy and institutional boundaries without obscuring limitations relevant to the recipient.

**Solution.** Connect audience-specific representations to versioned requirements, claims, production attempts, verification results and experience in use, as shown in Figure 7. Preserve assumptions and disagreements. Each view identifies simplifications and provides access to supporting material or relevant expertise. Changes to claims or evidence trigger review of affected views and understanding requirements. The technical, decision and public explanations in Figure 7 illustrate views tailored to engineers, decision-makers and community members. These views can share evidence about deployment limits while differing in mathematical detail and the value questions addressed.

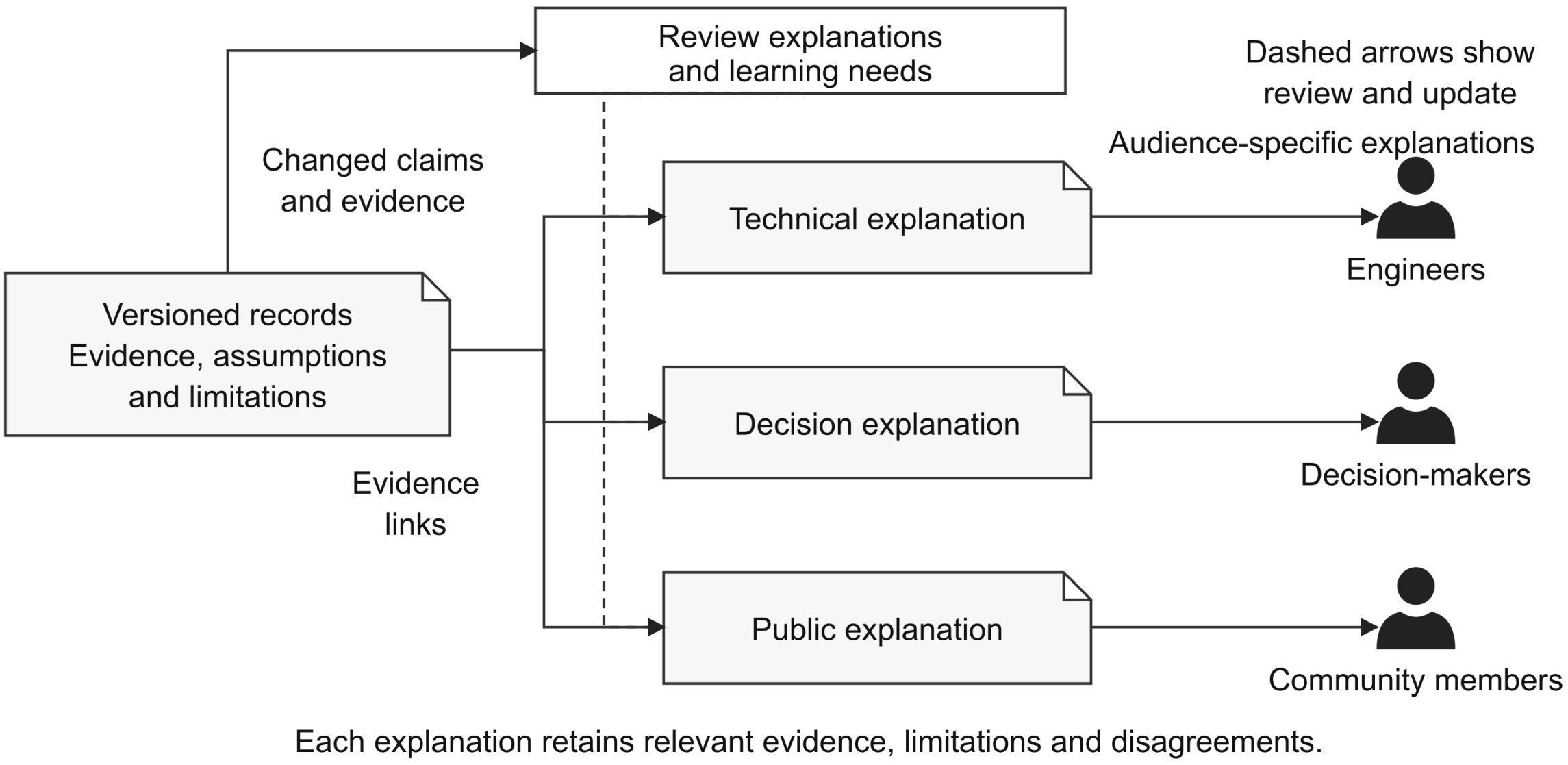

Figure 7. Shared evidence with audience-specific views.

**Consequences.**

**Benefits.**

- **Consistency.** Shared evidence reduces unintentional differences between accounts of the same claims and results.
- **Accessibility.** Audience-specific views address different knowledge and responsibilities.
- **Traceability.** Links to assumptions, evidence and expertise support examination of simplified accounts and identification of views requiring revision.

**Drawbacks.**

- **Maintenance cost.** Evidence links and dependent views require coordinated updates. A single view may suffice when recipient needs are similar.
- **Interpretive restriction.** A shared account can become an authorised interpretation. Independent accounts may be needed for adversarial scrutiny, with explicit provenance and comparison.

- **Access limitations.** Privacy restrictions can prevent recipients from inspecting underlying evidence, requiring clear disclosure of the limits of each view.

**Known uses.**

- **Intercom Fin and Copilot**[9]. A shared knowledge collection supplies AI-generated answers to customers through Fin and to support staff through Copilot. Content availability can be configured for each product, allowing different audiences to receive answers grounded in relevant shared material.
- **Google NotebookLM**[10]. Audio Overviews use uploaded sources and can be customised for an audience's expertise level and topics of interest. Different audiences can therefore receive differently focused AI-generated accounts based on the same source collection. These products illustrate shared-source and audience-tailoring mechanisms; preservation of disagreement, dependency-based revision and recipient understanding still require application-specific design and assessment.

### 6.5 Distributed expertise with tested handover (P5)

**Context.** The understanding required for AI-supported work may exceed any individual's capacity or available time. Organisations must maintain access to complementary expertise as membership and responsibilities change.

**Problem.** How can an organisation preserve usable understanding when responsibility moves between people or groups?

**Forces.**

- **Specialisation and coordination.** Complementary expertise reduces individual learning demands but creates dependencies between roles.
- **Continuity.** Transfer of responsibility requires usable understanding, whereas delivery of records alone does not establish the recipient's grasp.
- **Assessment feasibility.** Checking handover requires time and access to expertise while respecting privacy and professional responsibilities.

**Solution.** Allocate complementary understanding responsibilities across roles, identify dependencies and escalation routes, and maintain shared records, as shown in Figure 8. Test handover by asking the receiving group to locate and combine relevant expertise, explain scope, apply relevant knowledge and identify unresolved questions. Record understanding gaps and follow-up needs, then seek clarification and update the shared records and routes to expertise. Revisit the arrangement when members leave or assumptions change. Document delivery alone does not demonstrate continuity of collective understanding. Checking understanding supports, but does not authorise, transfer of responsibility. The physical, software and policy specialists in Figure 8 illustrate complementary responsibilities that the receiving team must connect.

---

9 https://www.intercom.com/help/en/articles/9459957-enable-or-disable-content-for-fin-and-copilot
10 https://blog.google/innovation-and-ai/products/notebooklm-update-october-2024/

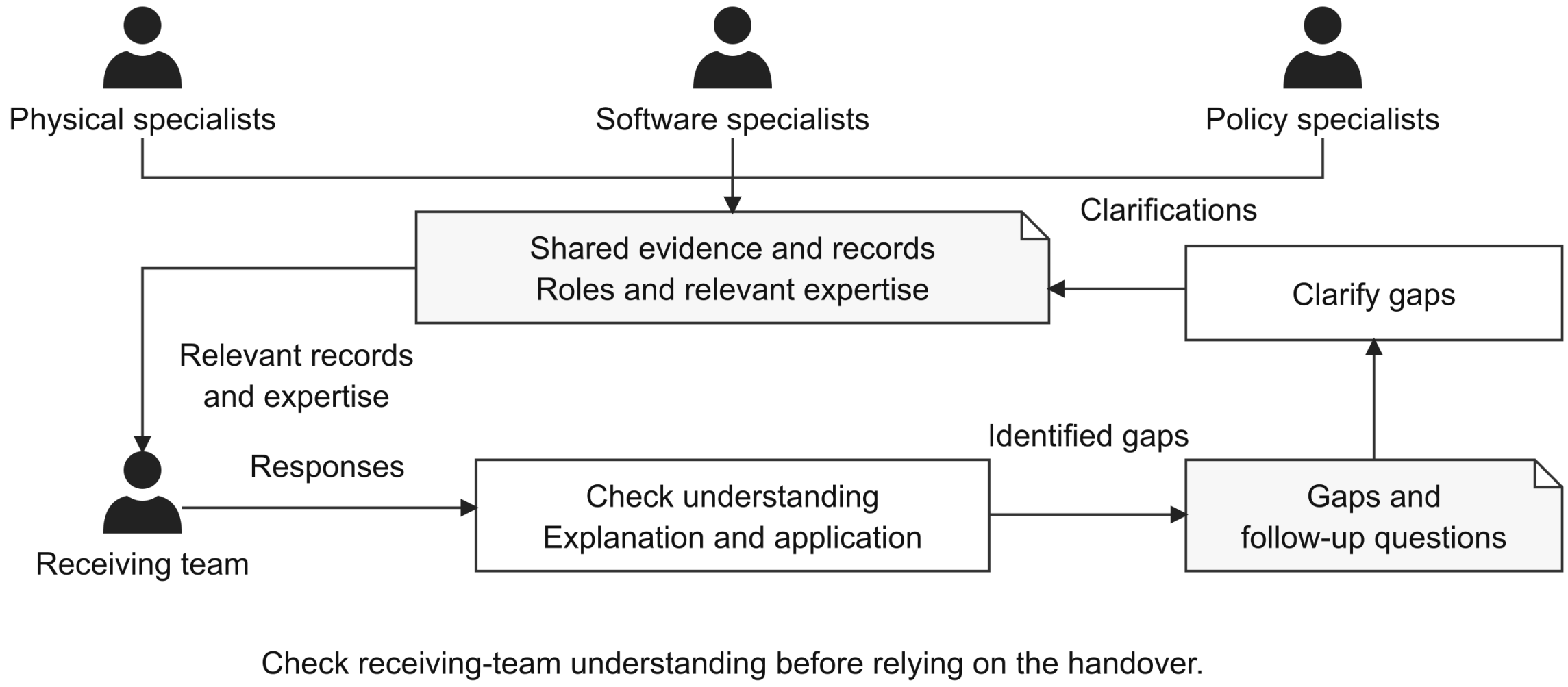

Figure 8. Distributed expertise with tested handover.

**Consequences.**

**Benefits.**

- **Efficiency.** Complementary responsibilities reduce duplication without requiring every participant to master every subject.
- **Continuity.** Shared records, expertise dependencies and escalation routes support access to knowledge when responsibilities change.
- **Visibility of gaps.** Receiver explanation and questioning can expose missing understanding before responsibility is transferred.

**Drawbacks.**

- **Coordination cost.** Maintaining shared records and conducting handover checks requires time. Individual mastery may be simpler for narrow responsibilities.
- **Expertise dependency.** Collective capability depends on available specialists and working relationships. Critical knowledge may require selective redundancy and reliable access to external expertise.
- **Assessment limitations.** A successful readback does not establish comprehensive understanding, and records of individual competence can become intrusive. Checks must remain proportionate to the responsibility.

**Known uses.**

- **Microsoft Copilot Studio**[11]. An AI agent can transfer a conversation to a human representative together with the conversation history and relevant variables. The receiving representative reviews the context before continuing. The application demonstrates context transfer, but the documentation does not describe testing the receiving human's understanding, so the example supports only the handover mechanism.
- **Intercom Fin.**[12] After Fin hands a conversation to a teammate, a workflow can add an AI-generated summary of the preceding interaction. The summary supports continuity by showing the receiving teammate what Fin attempted, but does not test the teammate's understanding.

[11] https://learn.microsoft.com/en-us/microsoft-copilot-studio/advanced-hand-off
[12] https://www.intercom.com/help/en/articles/15209304-ai-issue-summary-vs-ai-conversation-summary

- **Salesforce Agentforce.**[13] Conversational email escalation preserves the complete conversation history for the receiving human representative, who continues within the same email thread. This supports contextual continuity during handover, without establishing the receiving representative's understanding.

## 7 Evaluation

We examine the proposed approach through qualitative scenario analysis of two hypothetical use cases involving an AI-discovered energy technology and AI-supported public investment decisions. The evaluation examines whether the purpose classification helps identify distinct understanding requirements, how the components and their relationships in the reference architecture support those requirements, and under what conditions the architectural patterns provide useful design guidance. The cases exercise different purposes and design choices without representing the full range of AI applications. Each case connects a rationale to a recipient task, a design choice and a prospective check of understanding. The choices are hypothetical and the checks are proposed, not completed.

Each use case follows the same structure. Purposes and understanding requirements establish whose understanding is needed, of what and for what purpose. Architecture and pattern selection describe how the proposed design supports those requirements. Scenario-based assessment traces the design's response to an ordinary situation and an adverse condition, identifying support and gaps. Comparison with a simpler design examines whether the same requirements could be met with fewer distinct components or mechanisms, and identifies additional responsibilities and unresolved requirements. Findings and Limitations synthesises the findings across both cases and identifies the limits of the analysis.

Scenario analysis of architectural decisions and risks is informed by Kazman et al. (2000). The evaluation is an author-conducted analysis of proposed designs. The alternative designs are comparison baselines, not implemented systems tested in this study. The analysis does not constitute a stakeholder-based ATAM evaluation or measure changes in human understanding or operational performance. The findings are synthesised across both cases. Because the authors develop the framework and judge the designs, the analysis is susceptible to confirmation bias. Independent architectural review and assessment with intended recipients are needed to test the proposed benefits and identify omitted alternatives.

### 7.1 Energy technology understanding system

**Purposes and understanding requirements.** Consider an AI-supported investigation of degradation in an energy storage material. Assume automated tests have already established acceptable performance within a specified operating range, and additional human checking would not improve that acceptance decision. Two explanatory accounts remain consistent with the cycle and temperature records. Scientists want to understand which relationships are supported, why the accounts remain indistinguishable and what observation could separate them. This epistemic aim may justify inquiry even while operation continues unchanged. Operators need to apply the established limits and recognise when an observation calls for specialist investigation, an effective-action requirement. The scenario assumes an immediate operational schedule and a separate schedule for scientists' reconstruction and learning. These are conditions of the hypothetical case, not empirical findings about a material.

**Architecture and pattern selection.** Figure 9 follows the case's information and work: competing degradation accounts, design and checking of a discriminating trial, and revision of scientific and operational views. Shared records retain cycle and temperature observations, assumptions, competing predictions, claim status and approved operating limits. The scientific lead owns review of explanatory claims and proposed trials; the operating lead owns review of operational guidance. Learning requests can generate additional production and verification without automatically changing deployment. A finding that challenges an operating limit instead enters the ordinary operational response.

[13] https://help.salesforce.com/s/articleView?id=mktg.um_channel_email_human_rep_experience.htm&language=en_US&type=5

Parallel learning workflow (P1) addresses the stated difference in schedules: scientists reconstruct the competing predictions and trial design alongside continuing operation. *Shared evidence with audience-specific views* (P4) addresses the different scientific and operational requirements. A *separate explanatory model* (P2) is needed only if the available representations cannot support the required reasoning; it is unnecessary merely because two hypotheses exist. Stable operational goals do not call for P3. P5 applies only if responsibility later moves to another team, as considered in the short extension below.

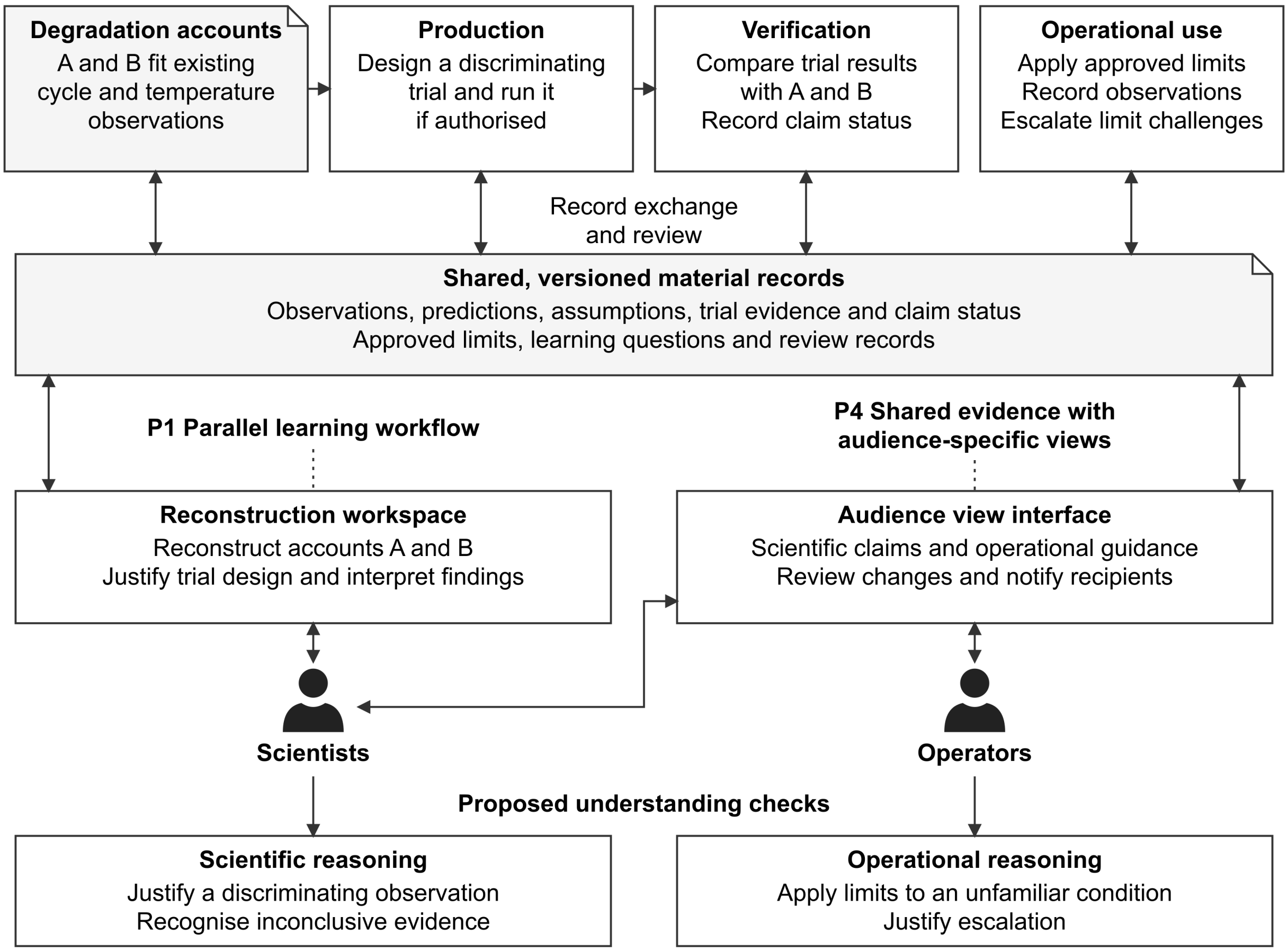

Figure 9. Energy case: discriminating inquiry and parallel learning linked to scientific and operational understanding. Continued operation is assumed to remain within already accepted limits unless new evidence challenges them.

**Scenario-based assessment.** Hold the accepted performance result and deployment decision constant. A scientist uses P1 to reconstruct the two accounts and explain why existing observations fail to distinguish them. The scientist and AI then propose a trial condition under which the accounts make different predictions. If a separate trial is authorised, production supports its performance and verification checks the evidence. The scientific lead records what the evidence supports and what remains unresolved. P4 carries that status into both the scientific account and the operating guidance. If further inquiry is not justified, the understanding target may be explicitly limited to grasping the alternatives and their evidential limits. The understanding requirement thus changes the allocation of work and evidence without requiring an improvement to the deployed result.

An adverse scenario arises when a later operational observation challenges a claim used in an explanation. The record identifies the claim, dependent views and review owners. The scientific lead reviews the inference, while the operating lead checks whether a limit or procedure is affected and initiates the applicable operational response. Recipients are notified of material revisions. Merely updating the shared repository would leave this coordination unspecified. Prospective assessment asks the scientist to justify a discriminating observation and recognise an inconclusive result, and asks an operator to interpret an unfamiliar condition using approved records and justify escalation. These tasks concern human grasp, separately from whether the AI's account is well supported.

**Comparison with a simpler design.** Figure 10 uses the same cycle records, hypotheses and operating limits with a scientific assistant and the team's existing work practices. It may meet the same requirements if people already own follow-up investigation, coordinated revision and assessment. Under the case's separate learning schedule, P1 makes reconstruction and its completion criteria explicit, while P4 links differently scoped accounts to the same claims and uncertainty. Neither additional workflow nor representation is needed when existing arrangements already provide these functions. The comparison concerns explicit support for recipient tasks and accountable follow-up, not the number of software boxes. Coordination cost, available time and assessment criteria still require application-specific decisions.

Consider removing these responsibilities while retaining the same assistant and records. Without an owner for follow-up investigation, an unresolved scientific question can remain documented without any decision about whether to pursue it. Without a review route connecting changed claims to dependent views, a revised scientific account need not reach operators. These are gaps in the specified design, not observed failures. Assigning the responsibilities within existing team practices can close the gaps without introducing separate services. The design implication is to specify responsibility and follow-up explicitly, rather than assume that an explanation interface provides them.

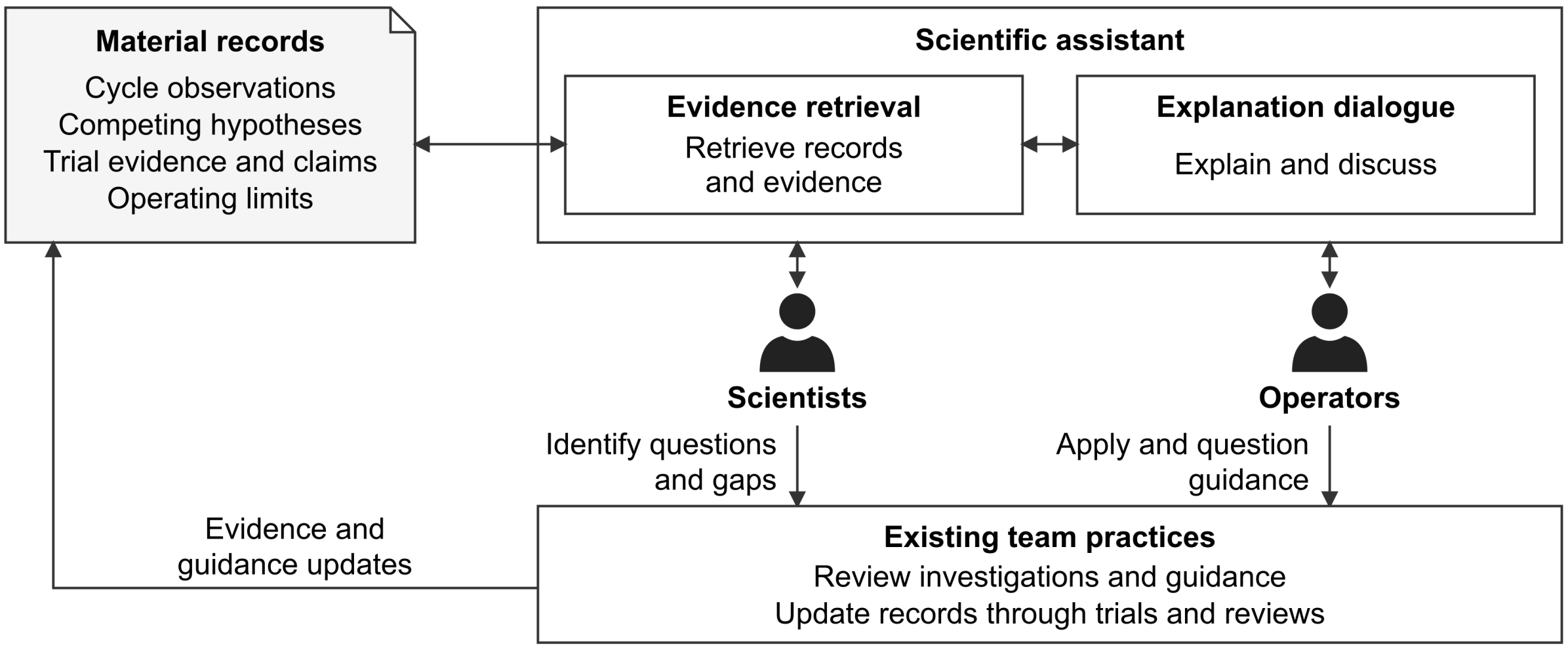

Figure 10. Simpler energy-case alternative: a scientific assistant supplies the same records and explanations, while investigation and understanding follow-up remain in existing team practices.

### 7.2 Public investment decision-support system

**Purposes and understanding requirements.** Consider a community comparing local energy portfolios with AI forecasts of energy output, land use and reliability. Assume the forecasts are adequate for the comparison and the immediate problem is not correcting them. Participants initially favour greater output but may reconsider land preservation after exploring consequences. Reflective agency requires being able to explain a retained or revised priority, including what was learnt about the trade-off. *Collective governance* additionally requires the decision-making body to connect evidence, dissent and its mandate to a legitimate choice. Curiosity about effects on a particular place can initiate this exploration without implying that participants must change their preferences or reach consensus.

**Architecture and pattern selection.** Figure 11 uses the same dotted-link notation for pattern applications. Solid arrows represent information exchange and review routes, with operational use subject to authorisation by the decision-making body. The architecture supports two concrete goal settings: prioritising energy output and giving greater weight to land preservation. Production generates alternative portfolios using the same forecasting model and assumptions. Verification checks evidence and applicable constraints for each portfolio. Shared records retain both goal settings, technical uncertainty, reasons and disagreement. Learning and exploration supports a goal exploration workspace and an audience view interface, with P3 and P4 identifying the patterns applied to those components. The decision-making body reviews proposed revisions against its mandate. The use component applies only authorised goals and approved portfolios to operational planning. Understanding supports the formation and examination of preferences, while decision authority remains an institutional matter.

*Branch and compare goals* (P3) is selected because the participants' priorities may change through exploration and must remain distinct from approved commitments. *Shared evidence with audience-specific views* (P4) is selected because participants and technical reviewers need different representations of the same forecasts and uncertainty. The case specifies neither a separate learning schedule nor a need for an additional explanatory model, so P1 and P2 are not assumed. No transfer of responsibility is involved, so P5 is unnecessary. These exclusions follow from the scenario requirements rather than a requirement to use every pattern.

**Scenario-based assessment.** Participants compare an output-prioritising portfolio with a branch that gives greater weight to land preservation. P3 retains the original goal and alternatives for comparison, while verification checks the resulting portfolios against the relevant constraints. P4 presents their implications for output, land use and reliability with the relevant uncertainty qualifications preserved in both views. A participant may favour the branch, retain the original priority or reject both. Reasons and remaining disagreements are recorded before the decision-making body authorises a portfolio or requests further exploration. The forecasting model and assumptions can remain unchanged throughout, although different portfolios may have different predicted outcomes. The relevant development is the participants' understanding of what they value and why. A prospective assessment asks participants to explain a trade-off and its uncertainty under an unfamiliar alternative and fairly describe a reason for an opposing choice. Agreement with the adopted portfolio is not the attainment criterion.

An adverse scenario arises when a public view omits uncertainty retained in the technical account. A named explanation owner reviews both against the current forecasts and corrects or withdraws a materially misleading view. The decision-making body determines whether the correction requires reconsideration before commitment. Participants must also be able to question the framing and introduce alternatives. A prospective recipient task would assess understanding of the qualification, rather than infer understanding from distribution of a corrected document. These responsibilities are required by the selected understanding purposes; neither shared storage nor P3 alone supplies them.

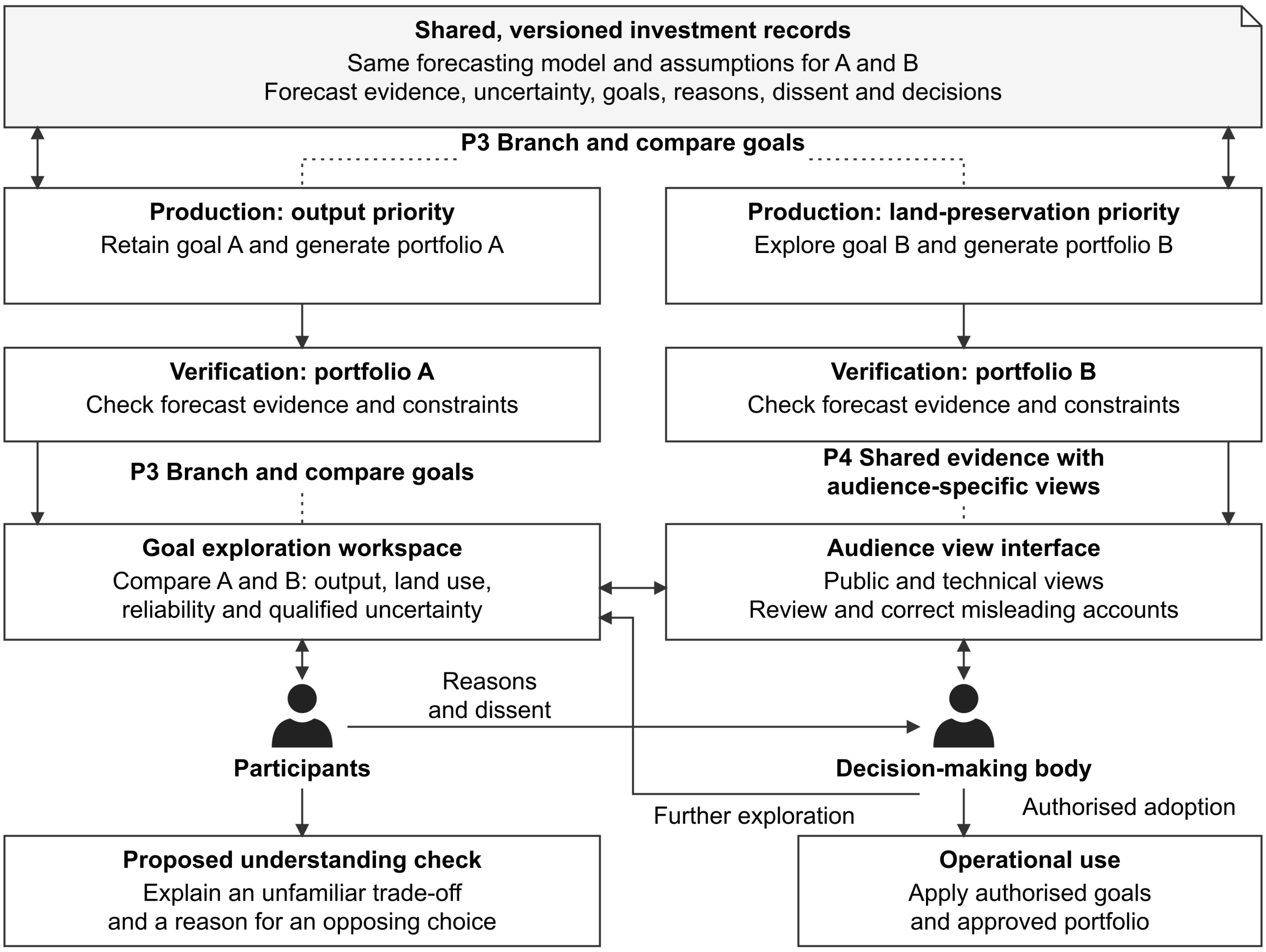

Figure 11. Investment case: compare output and land-preservation priorities, examine shared uncertainty through audience-specific views, and separate reflective preference formation from authorised adoption.

**Comparison with a simpler design.** Figure 12 uses an interactive planning assistant, versioned portfolios and an approval process. Such a design can already preserve alternatives and separate exploration from adoption. It qualifies as understanding-first if explicit recipient requirements guide the comparisons, the treatment of uncertainty and follow-up on participants' reasoning. P3 and P4 describe useful choices within either implementation; they do not require extra services. Where a common view is adequate, separate representations add maintenance cost without a corresponding benefit. The comparison therefore asks who ensures that participants can reason about the alternatives, how misleading simplifications are corrected and how disagreement reaches the authorised decision process. Legitimacy additionally depends on representation and decision rules beyond the software architecture.

The same comparison exposes two distinct omissions. Retaining alternative portfolios without separating exploration from authorisation leaves the status of a proposed goal revision unspecified. Publishing public and technical views without a review owner leaves responsibility for inconsistent uncertainty qualifications unspecified. P3 and P4 describe arrangements addressing these omissions, but existing versioning and review processes can provide the same functions. The scenario therefore supports conditional pattern selection and explicit decision authority, not a claim that a more elaborate architecture improves understanding.

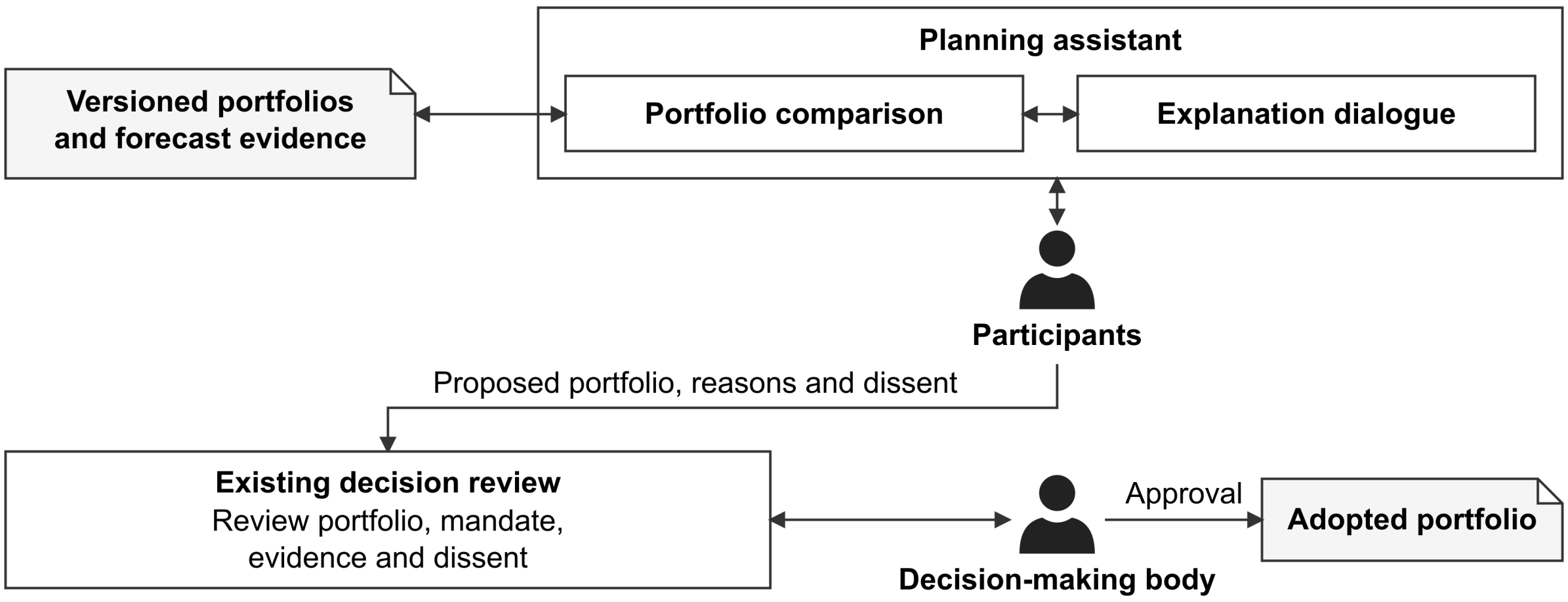

Figure 12. Simpler investment-case alternative: existing planning, versioning and approval facilities may meet the same understanding requirements when comparison and follow-up responsibilities are explicit.

### 7.3 Findings and Limitations

The two illustrative cases show how the purpose classification distinguishes understanding requirements for different recipients. In the energy case, scientists need to compare competing explanations, while operators need to understand operating limits and uncertainty. In the investment case, participants need to understand priorities and trade-offs, while the decision-making body needs to account for evidence, disagreement and decision authority. These distinctions guide the choice of representations, activities and assessment tasks. The cases demonstrate application of the classification, but do not establish that the four rationales cover every purpose of understanding. In the ordinary scenarios, the energy case examines understanding-driven investigation with the original acceptance decision held constant, while the investment case examines reflective preference formation using unchanged forecasting assumptions. These comparisons make the purposes beyond error correction consequential to design.

The architecture connects requirements to records, activities and responsibilities. In the energy case, scientific and operating leads review different consequences of an evidential change. In the investment case, an explanation owner reviews public qualifications and the decision-making body controls commitment. The scenarios specify these roles to expose what shared records alone leave unresolved. Application-specific assessment criteria, notification arrangements and institutional authority still need to be agreed. The scenario analysis does not establish the effectiveness of these arrangements.

The pattern coverage is selective. P1 and P4 are exercised in the energy scenario, and P3 and P4 in the investment scenario. P2 remains conditional on the need for a separate explanatory representation. A short extension illustrates P5: if a new team assumes responsibility for the material, it uses the records and relevant specialists to explain approved limits, identify unresolved degradation claims and justify an escalation route before handover is relied upon. Gaps trigger clarification and another check. This is a prospective handover walkthrough, not an observed transfer. Figures 4–8 illustrate all five candidate mechanisms. The two main scenarios examine selected combinations rather than validate the whole catalogue.

Across both comparisons, existing explanation, evidence and decision-support mechanisms may already meet some understanding requirements. The proposed approach provides a basis for specifying recipient needs and assigning responsibilities for investigation, explanation and revision. Additional representations or workflows are justified only when those requirements remain unmet. Otherwise, additional representations or workflows may introduce avoidable maintenance and coordination costs. A simpler implementation can satisfy the same understanding commitments. The comparisons identify design conditions and trade-offs, but do not establish superiority over existing systems.

The hypothetical scenarios illustrate how understanding purposes guide architectural choices and expose obligations that a component inventory alone does not settle. They do not measure human understanding or demonstrate effectiveness. Independent review and empirical studies should compare implementations using the same recipient requirements and matched tasks, examining reasoning about unfamiliar cases, retention where relevant, coordination cost and renewal after evidence changes. The short P5 extension likewise requires evaluation with an actual receiving team.

## 8 Conclusion

Understanding-first AI engineering makes purpose-specific human understanding an explicit outcome that can shape requirements, design and operation. Error correction remains important, but its removal does not eliminate reflective agency, collective governance or curiosity as reasons to understand. The four rationales, reference architecture and five candidate patterns connect those reasons to engineering choices. The hypothetical cases show how inquiry and participation can change with technical acceptance held constant, and how people can reconsider priorities without correcting forecasts. Their effectiveness remains to be tested. A continuing research agenda is to assess recipient-relative grasp without rewarding fluent imitation, allocate work and resources to independently valued understanding, and sustain collective understanding as evidence and responsibilities change.